\documentclass[10pt,twocolumn,letterpaper]{article}

\usepackage{wacv}
\usepackage{times}
\usepackage{epsfig}
\usepackage{graphicx}
\usepackage{amsmath}
\usepackage{amssymb}

\wacvfinalcopy 

\def\wacvPaperID{1157} 

\ifwacvfinal\fi
\begin{document}

\title{
Analysis and a Solution of Momentarily Missed Detection\\ for Anchor-based Object Detectors
}

\author{
    Yusuke Hosoya${}^\text{1}$
    \quad
    Masanori Suganuma${}^\text{1,2}$
    \quad
    Takayuki Okatani${}^\text{1,2}$\\
    ${}^\text{1}$Graduate School of Information Sciences, Tohoku University
    \quad
    ${}^\text{2}$RIKEN Center for AIP\\
    {\tt\small \{yhosoya,suganuma,okatani\}@vision.is.tohoku.ac.jp}
}

\newtheorem{hyp}{Hypothesis}

\maketitle
\ifwacvfinal\thispagestyle{empty}\fi

\begin{abstract}
    The employment of convolutional neural networks has led to significant performance improvement on the task of object detection. However, when applying existing detectors to continuous frames in a video, we often encounter momentary miss-detection of objects, that is, objects are undetected exceptionally at a few frames, although they are correctly detected at all other frames. In this paper, we analyze the mechanism of how such miss-detection occurs. For the most popular class of detectors that are based on anchor boxes, we show the followings: i) besides apparent causes such as motion blur, occlusions, background clutters, etc., the majority of remaining miss-detection can be explained by an improper behavior of the detectors at boundaries of the anchor boxes; and ii) this can be rectified by improving the way of choosing positive samples from candidate anchor boxes when training the detectors.
\end{abstract}


\color{black}

\section{Introduction}\label{sec:intro}

\begin{figure}[t]
    \begin{center}
        \includegraphics[keepaspectratio, width=\linewidth]{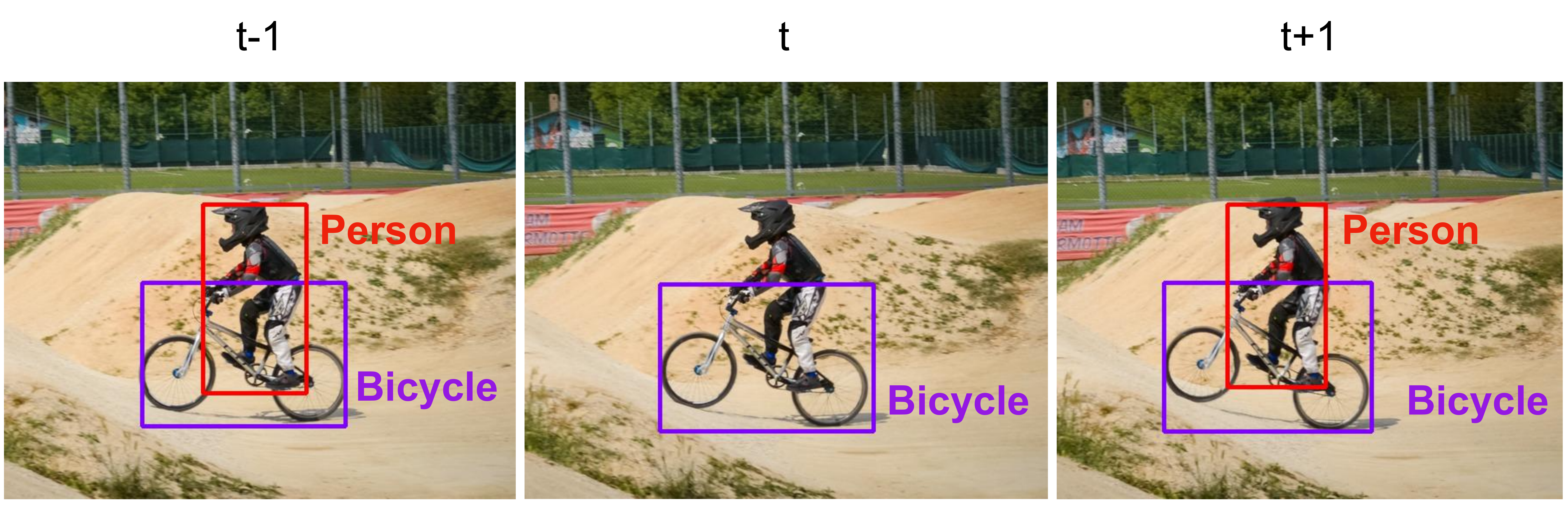}\\
    \end{center}
    \caption{An example of momentarily missed detection of an object. The person riding a bicycle is continuously detected in a sequence of frames but is not detected at the frame in the middle. Unlike those in Fig.~\ref{fig:external_factor}, there is no apparent cause explaining the miss-detection.}
    \label{fig:mmd_example}
\end{figure}

\begin{figure}[t]
    \begin{center}
        \includegraphics[keepaspectratio, width=\linewidth]{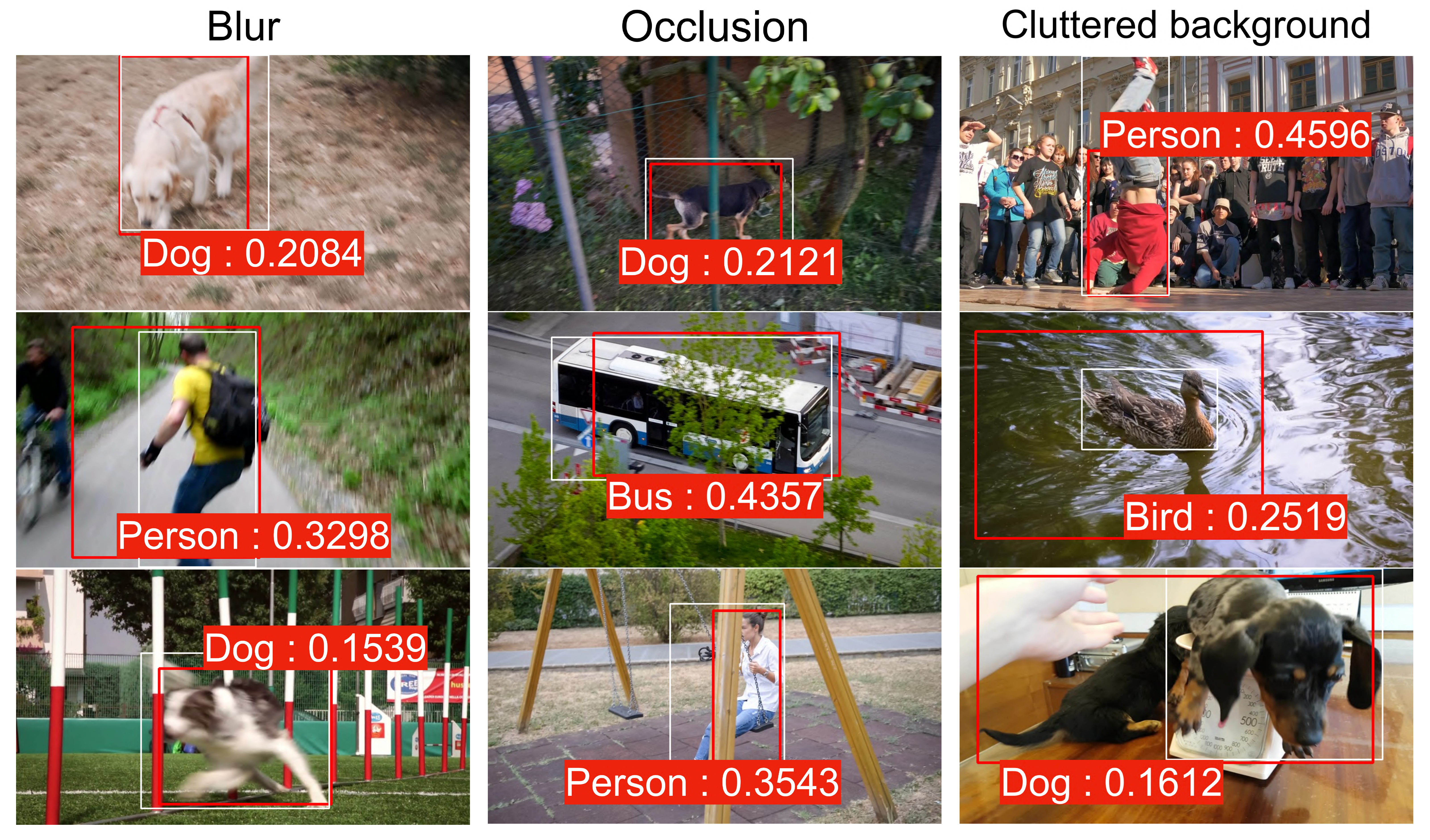}\\
    \end{center}
    \caption{Examples of the miss-detection cases for which we can see apparent causes, i.e., {\em external} factors. The white boxes indicate the ground-truth boxes and the red ones with a score indicate the predictions with the highest scores amongst those with IOU$>0.5$. From left to right, the miss-detection is considered to occur due to motion blur, occlusion, and cluttered background. }
    \label{fig:external_factor}
\end{figure}

Detecting objects in an image is a fundamental problem of computer vision. The employment of convolutional neural networks (CNNs) has led to significant performance improvement on the task in recent years \cite{FasterRCNN,SSD,yolov2,RetinaNet,RefineDet,M2Det,CornerNet}, and the accuracy seemingly comes close to the upper bound of the task. However, when applying them to a video, there often emerge cases where an object is momentarily miss-detected, i.e., undetected exceptionally at a few frames, although it is successfully detected for all other frames
\cite{DetStability,Flowguided,STMN}; examples are shown in Figs.~\ref{fig:mmd_example} and \ref{fig:external_factor}.

In this paper, we attempt to understand the mechanisms of how such miss-detection, which we will call momentarily missed detection (MMD),  occurs. While in some cases it is apparent what causes MMD, e.g., motion blur and occlusion, as shown in Fig.~\ref{fig:external_factor},  in others it is unclear, as in the example of Fig.~\ref{fig:mmd_example}. We call the former {\em external} factors and the latter {\em internal} factors. 

To the authors' knowledge, there is no study in the literature that shows detailed analyses as to how many out of such miss-detection cases are attributable to external factors and internal factors, and what mechanisms are behind the internal factors. This may be reasonable, since MMD by definition occurs only rarely and thus has only small impact on detection accuracy that is usually measured by mAP (mean average precision).

In this paper, aiming to answer these questions, we conduct a series of analyses on the class of detectors that use anchor boxes, which is currently the most popular approach. In our experiments, we analyzed how SSD \cite{SSD} and M2Det \cite{M2Det} work on 73 videos of diverse scenes contained in the DAVIS dataset \cite{DAVIS}. We then found that MMD occurs at a few percent of about 8,000 frames in these videos, for which we will show the followings:
\begin{itemize}
    \item About 60 to 70\% of the MMD cases are attributable to external factors, such as motion blur, occlusion, background clutter, etc.
    \item About 20 to 30\% of the MMD cases, for which there is no apparent cause like the external factors, can be explained by an improper behavior of the detectors at the boundaries of anchor boxes. Specifically, when an object moves in a video, its predicted score can drop considerably at the very instant when the optimal anchor for the object is switched from one to its neighbor.
    \item This behavior can be rectified by improving the method of choosing positive samples out of candidate anchors when training the detectors. This prevents the occurrence of MMD in most of the above cases without sacrificing overall detection accuracy; it also contributes to reduce MMD cases caused by the external factors.
\end{itemize}

As MMD occurs only rarely, its removal contributes only a little to improving detection accuracy that can be measured by mAP. Then, a practical value of our study, in addition to deeper understanding of how the detectors work,  is as follows. The major application of object detection is arguably tracking moving objects in a video. To use a CNN-based detector for this purpose, we will first apply it to each video frame independently and then associate the detection results over the frames. Before the emergence of powerful CNN-based detectors, the second step of data association was relatively important, as the first step can only be performed with limited accuracy. The recent CNN-based detectors achieve much higher accuracy, which is sometimes nearly perfect. This lowers the relative importance of the second step, except for some hard cases, e.g., when many moving objects overlap with each other. If we can  reduce the occurrence of MMD toward zero, we may be able to further accelerate this trend.


{

}

\section{Related Work}\label{related_work}

{
}
\color{black}


In recent studies, object detection is posed as a regression problem, in which the geometry of an object bounding box in an image is predicted.
There exist a few approaches of formulating this regression problem.
The major approach is to use {\em anchor boxes}, which are default bounding boxes with several predefined sizes and aspect ratios that are positioned at each grid cell of a feature map. The problem is then formulated as prediction of offsets to the true bounding box from its closest anchors, making it easier to deal with bounding boxes which usually have high degrees of freedom. 
The methods based on this  formulation are further divided into two categories.
One uses multiple feature maps in different resolutions \cite{SSD, DSSD, FPN, RetinaNet, RefineDet, M2Det} and the other use a single feature map \cite{FasterRCNN, yolov2}. These two strategies are compared in \cite{FPN, GS_Face}. 

{


}

For these methods, in this paper, we analyze momentary miss-detection that is often observed when applying them to a video sequence. To the authors' knowledge, there are few studies on this issue in the literature. 
An exception we are aware of is the study of Zhang \etal \cite{DetStability}, in which considering the situation where a single-image object detector is applied to a continuous sequence of a video, they propose a new measure evaluating temporal (in)consistency of detection results and show that there is a trade-off between accuracy and stability. Our results in this paper do not disagree with theirs but show that we can improve the trade-off. It should also be noted that there are a number of studies that consider video object tracking, such as \cite{Flowguided} and \cite{STMN}, in which the input is a set of multiple frames and the output is the trajectories of objects. 



Recently, several methods have been proposed that do not use anchor boxes, such as
CornerNet \cite{CornerNet} and CenterNet \cite{CenterNet}. 
Instead of predicting the offset from anchor boxes, they directly predict key points associated with bounding boxes, such as their top-left and bottom-right corners. Tian \etal \cite{FCOS} propose a method that directly predicts the location of an object at each grid point of a feature map, which is inspired by FCN-based architectures \cite{DenseBox}. Although these approaches seem promising, it is too early to say that they will fully replace anchor-based methods.  
It is noteworthy that several researchers have proposed to use additional components to improve the anchor-based methods. Yang \etal \cite{Metaanchor} propose a method that dynamically generates appropriate anchor boxes from the input by using an additional network called the anchor function generator.
Zhu \etal \cite{FeatureSelective} propose to choose the best one of multi-scale feature maps for each input and predict bounding boxes at its grid points, which is integrated with predictions from the standard anchor-based method.

{
}

\color{black}
\section{Momentarily Missed Detection}\label{sec:mmd}

We are interested in cases where detection of an object fails at a  particular frame of a sequence and is successful for all the other frames in the sequence. We call such miss-detection {\em momentarily missed detection} (MMD). 

\subsection{Definition}\label{sec:3.1}\label{sec:mmd.def}

To conduct systematic analyses, we declare the frame $t$ to be a MMD frame if it satisfies the following conditions:
\begin{subequations}
\label{eq:mmd_condition}
\begin{align}
    &{p_{t-1}^{c}} \geq \gamma_{min} \ \ \mbox{and} \ \ \ {p_{t+1}^{c}} \geq \gamma_{min},\\
   &p_{t}^{c} / p_{t-1}^{c} \leq \gamma_{ratio},\\
   &{p_{t}^{c}} < \gamma_{max},
\end{align}
\end{subequations}
where $p_t^c$ is the score of the object of class $c$ that is given by our detector for frame $t$. Rigorously, we select  the anchor box providing the largest score from those having IOU$>0.5$ with its ground truth  box; $p_t^c$ is the score for the anchor. The first condition (\ref{eq:mmd_condition}a) ensures that the last and next frames (i.e., $t-1$ and $t+1$) provide sufficiently high detection score. The second one (\ref{eq:mmd_condition}b) requires that detection score drop at this frame $t$ from the last frame $t-1$, at least to a certain extent. 
\color{black}
The last one (\ref{eq:mmd_condition}c) eliminates frames at which the detection score is very high although the first two conditions are met. 

When a detector detects objects successfully (except for MMD frames), the score $p_t^c$ defined as above is usually greater than $0.7$ (for any $t$), and thus (\ref{eq:mmd_condition}c) with $\gamma_{max}$ having a lower value with a certain gap suffices to find most of the MMD frames of our interest. Thus, we set $\gamma_{max}=0.6$ in our experiments. However, there are additional cases where the score is continuously close to (but greater than) a natural threshold $0.5$ and exhibits an exceptionally lower score at a single particular frame $t$. For completeness, we want to extract such frames as well, employing (\ref{eq:mmd_condition}a) and (\ref{eq:mmd_condition}b); we set $\gamma_{min}=0.5$ and $\gamma_{ratio}=0.9$ in our experiments.

{
}

\subsection{Factors Causing MMD}\label{sec:mmd.pf}

\subsubsection{External Factors}\label{sec:mmd.pf.ext}

We now consider why MMD occurs, i.e., why an object that is continuously detected suddenly ceases to be detected at a particular frame. The most likely cause will be bad imaging conditions that momentarily emerge at that frame, such as motion blur, illumination changes, occlusion by other objects, cluttered background, etc. Figure~\ref{fig:external_factor} shows typical examples, which are selected from those automatically selected by the conditions (\ref{eq:mmd_condition}).
It should be noted that several solutions to these cases are proposed in \cite{RTR, UnderMB, ResMap}.
We refer to these causes as {\em external} factors. 

\subsubsection{Anchor Boundary}\label{sec:mmd.pf.anchor}
\begin{figure*}[t]
    \begin{center}
        \scalebox{0.9}{
            \includegraphics[keepaspectratio, width=\linewidth]{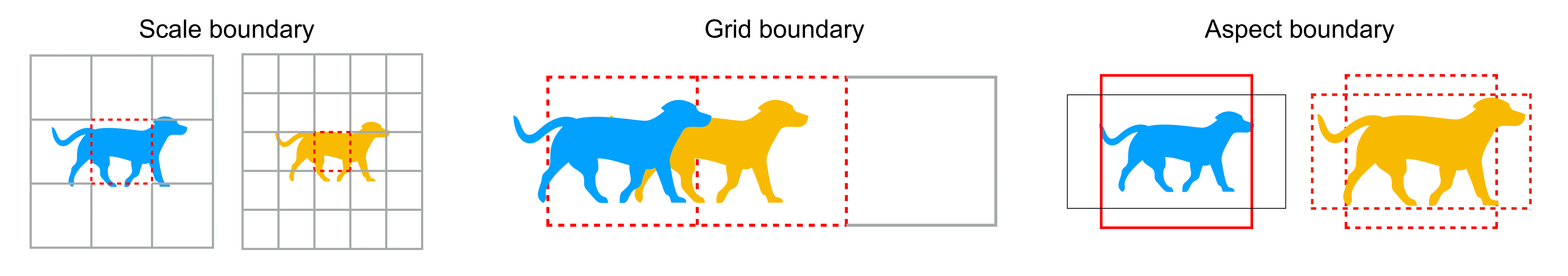}
        }
    \end{center}
    \caption{Three types of boundaries between neighboring anchor boxes. }
    \label{fig:be}
\end{figure*}

However, there are cases where we cannot find such apparent causes, as shown in Fig.~\ref{fig:mmd_example}. What causes MMD in such cases? We conjecture it is attributable to (suboptimal) use of anchor boxes. 

We are considering detectors that use anchor boxes. They are created at each grid point of a feature map with a number of sizes and aspect ratios, aiming at easing regression of bounding boxes with high degrees of freedom. Thus, there are usually a large number of anchor boxes with various locations, sizes, and aspect ratios. 
However, they are only sparsely sampled in these parameter space.
Thus, when a moving object continuously changes its location, size, and shape in a video, the optimal anchor that is responsible for its detection will be switched from one to its neighbor, as shown in  Fig.~\ref{fig:be}. It is noteworthy that as the fewer the anchors are, the faster the overall computation is, it is not reasonable to use an excessive number of anchors.  

{
}

Based on this structure of the detectors, we introduce the following hypothesis as to how MMD emerges that cannot be explained by the external factors:

\begin{hyp}
MMD can emereg at the boundaries of two neighboring anchors having either different scales, locations, or aspect ratios. 
\end{hyp}

We will experimentally validate this hypothesis. To do so, we first propose a method for analyzing the behavior of detectors at around anchor boundaries. 

\color{blue}


\color{black}

\section{Analysis and a Solution}\label{sec:ASmmd}

\subsection{Understanding Behaviors of a Detector}\label{sec:ASmmd.visualize}

\subsubsection{Image Warp Simulating Object Motion around Anchor Boundaries}\label{sec:ASmmd.visualize.synthesis}

Suppose we find MMD that occurs at an image of a sequence and wish to analyze how our detector behaves on that image and its neighbors in the sequence. To do this, we simulate the image motion of the object by warping the image at which the MMD occurs and examining detector behaviors on the warped images. 

\begin{figure}[t]
    \begin{center}
        \includegraphics[keepaspectratio, width=\linewidth]{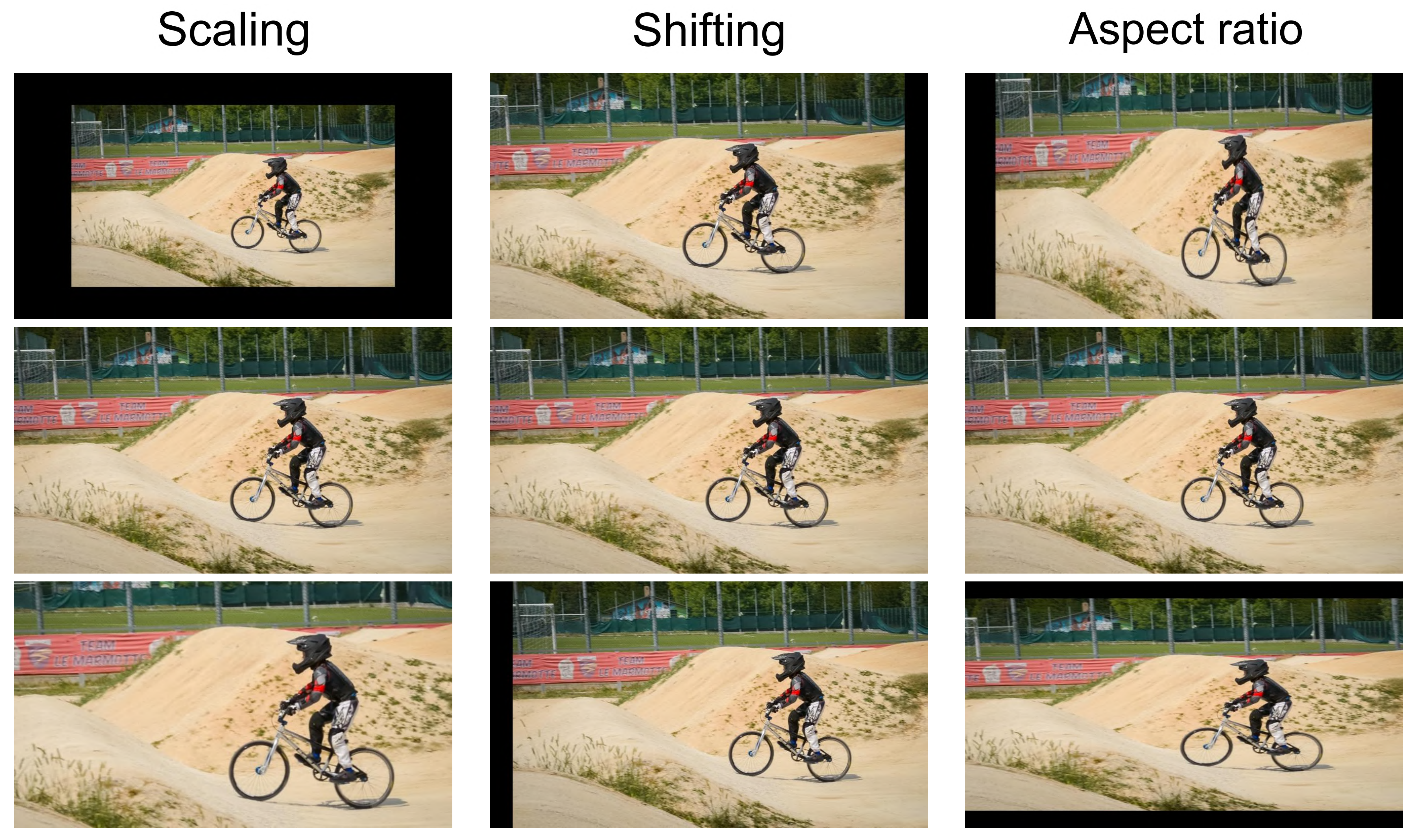}\\
    \end{center}
    \caption{Examples of the transformed images by scaling, horizontal shift, and aspect ratio (from left to right). The top and bottom rows show the transformed images at $n=\pm 15$. If an object is miss-detected in the original image because the object size is on the boundary of anchors, then these transformed images will break the balance, making detection of the object back to normal. }
    \label{fig:transforms}
\end{figure}

To be specific, we first apply a series of geometric warp to the image, generating image sequences, in each of which the object appears in either different sizes, positions, or aspect ratios. Then, the optimal anchor for the object will be switched from one to another in the images in each sequence, as shown in Fig.~\ref{fig:transforms}. Next, we run the detector on each image sequence and examine its outputs. {\em If MMD does occur at a boundary of anchors, we should be able to observe a decrease in the predicted score for the object at around the switch of the optimal anchor.}

To consider the three types of anchor boundaries shown in Fig.~\ref{fig:be}, we consider scaling, horizontal shift, and change in aspect ratio for the image warping. Their details are as follows:

\smallskip
\noindent
{\bf Scaling:} The image is enlarged and shrunk by a factor of $1.02^n$ and $0.98^n$ respectively, for $n=1,\ldots,29$, yielding $59$ images including the original.

\smallskip
\noindent
{\bf Shifting:} We consider only horizontal shift. The image is shifted in the $x$ axis by $3n$ pixels, for $n=-29,\ldots,29$, yielding $59$ images in total. 

\smallskip
\noindent
{\bf Aspect ratio:} The image is enlarged and shrunk in either of $x$ or $y$ axis by a factor of $1.01^n$ and $0.99^n$, respectively, for $n=1,\ldots,29$, yielding $59$ images in total for each axis.

\subsubsection{Case Studies}\label{sec:ASmmd.visualize.case_s}

\begin{figure}[t]
    \begin{center}
        \includegraphics[keepaspectratio, width=\linewidth]{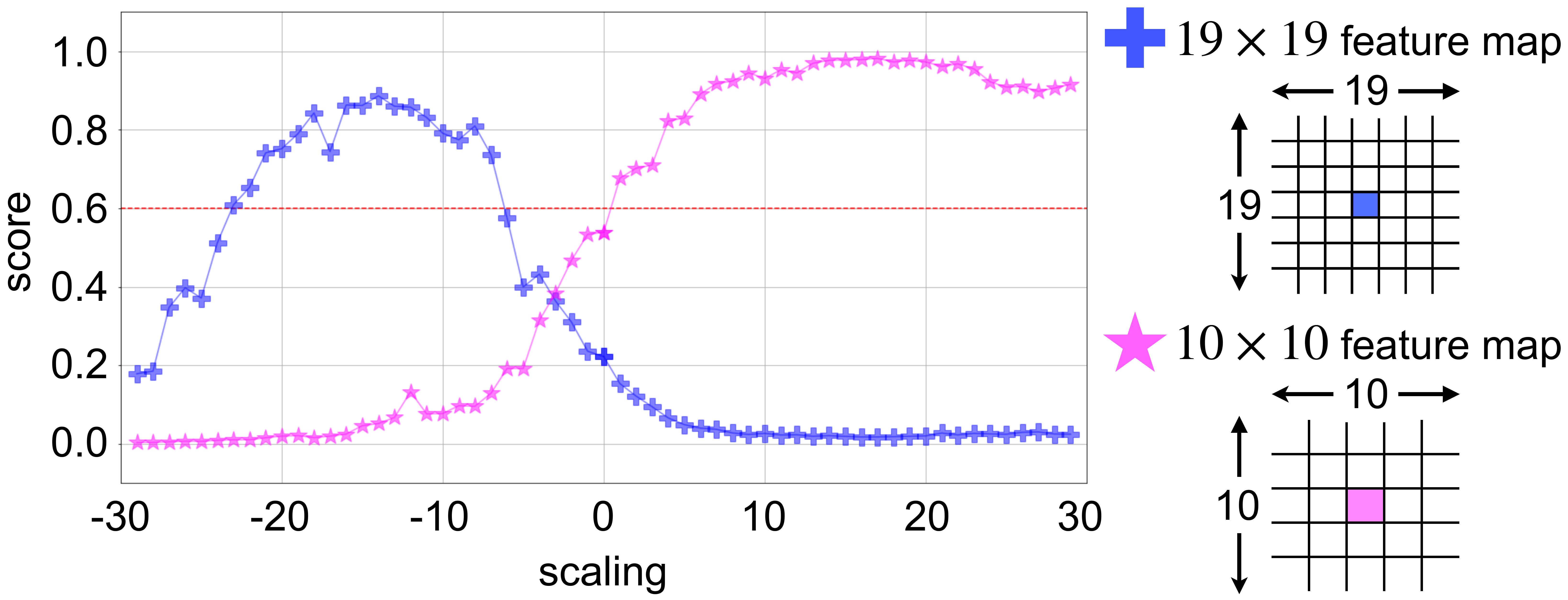}\\
    \end{center}
    \caption{An example of how a detector (SSD) behaves around a boundary of two anchors with different sizes (scales). The crosses and stars indicate the score of two neighboring anchors, one from $19\times 19$ and the other from $10\times 10$ feature maps, respectively, for a sequence of images that are obtained by scaling the original image; the horizontal axis indicates the scaling factor, e.g., $1.02^n$.} 
    \label{fig:scale_boundary}
\end{figure}

How does a detector behave around an anchor boundary? A typical behavior is shown in Fig.~\ref{fig:scale_boundary}. The plot shows how the score for an object varies with respect to its size change, which is simulated by scaling the image. The blue crosses indicate the scores predicted for an anchor created on the $19\times 19$ feature map, whereas the magenta stars indicate those for an anchor on the $10\times 10$ feature map. 
It is seen from Fig.~\ref{fig:scale_boundary} that the former gives higher scores in the negative $(n<0)$ region (i.e., smaller object size), whereas the latter gives higher scores in the positive $(n>0)$ region (i.e., larger object size), indicating that the optimal anchor is switched at around the original object size ($n=0$). It is also seen that the detection score becomes lower than it should be at around the original size $(n=0)$. Ideally, the score should be kept high throughout a range centered at $n=0$, resulting in stable detection of the object. The lower scores at around $n=0$ provide a direct explanation as to why MMD occurs at the original image.

\begin{figure}[t]
    \begin{center}
        \includegraphics[keepaspectratio, width=\linewidth]{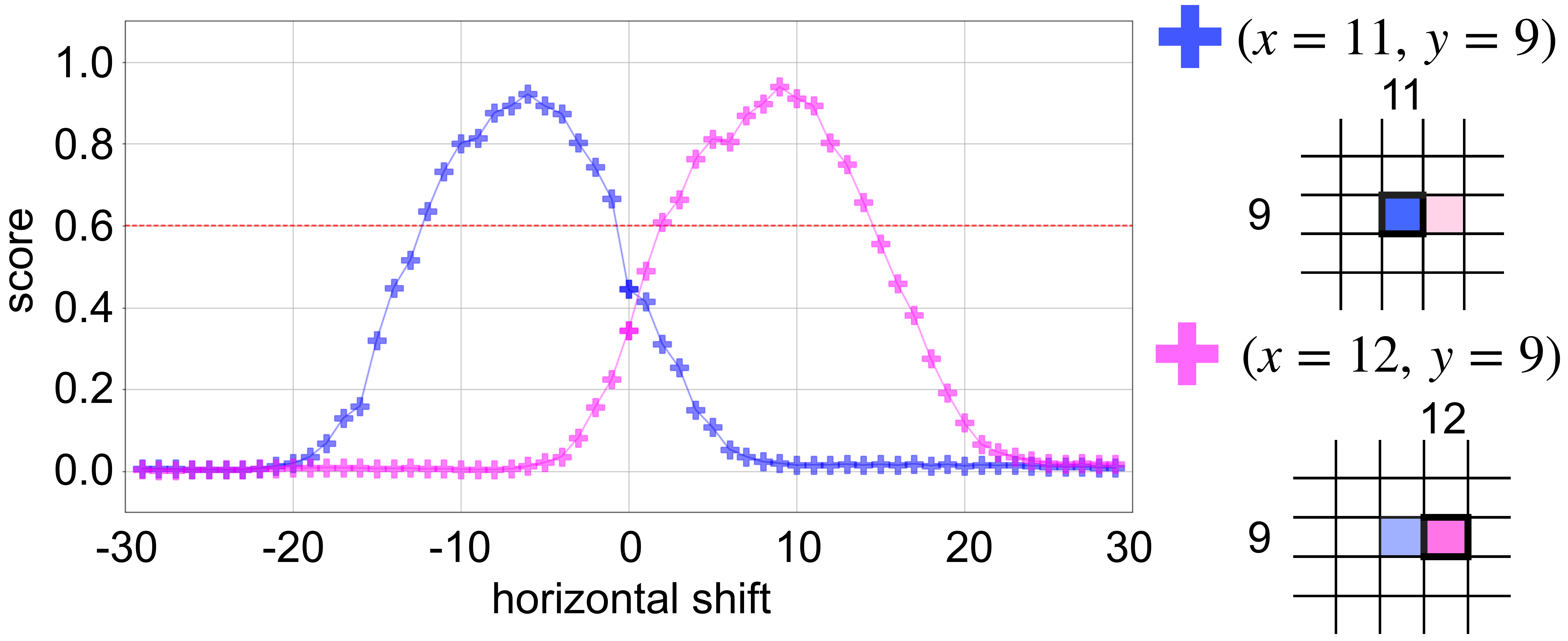}\\
    \end{center}
    \caption{An example of how a detector (SSD) behaves around a boundary of two anchors on different grid points. The blue and magenta crosses indicate the score for two neighboring anchors, one from the grid point $(x=11, y=9)$ and the other from $(x=12, y=9)$ in the $19\times19$ feature map, respectively.}
    \label{fig:grid_boundary}
\end{figure}

As shown in Fig.~\ref{fig:grid_boundary}, a similar behavior is observed for boundaries of anchors defined on different grid points. The blue crosses and the magenta crosses show detection scores for two neighboring anchors, $(x,y)=(11,9)$ and $(12,9)$, respectively, which are created on the same $19\times 19$ feature map. The former gives higher scores in the negative region (i.e., shifting the image to the left) and the latter gives higher scores in the opposite region (shifting to the right). The optimal anchor is switched at around the center, where the scores for the two anchors are both lower than they should be. This explains why MMD occurs at the original image. 


We will show the results of experiments conducted on a large number of images to validate Hypothesis 1 in Sec.~\ref{sec:result}.

\subsection{Improving Detector Behavior around Anchor Boundaries}\label{sec:ASmmd.improve}

Assuming Hypothesis 1 to be true, how can we improve the behavior of detectors around anchor boundaries? We conjecture that the drop of detection score on anchor boundaries, as seen in Figs.~\ref{fig:scale_boundary} and \ref{fig:grid_boundary}, is attributable to sub-optimal methods for selecting positive samples. We then present an improved method. 


\subsubsection{Conventional Method for Positive Sample Selection}\label{sec:ASmmd.improve.conventional}

To achieve high recall, many recent detectors has a common design that a CNN first generates a number of candidate boxes for a single object, and they are all discarded but a single box in a post filtering process, such as non-maximum suppression and center-ness scores \cite{FCOS}. Thus, when training the CNN, multiple anchor boxes are selected as positive samples; that is, we train the CNN to declare the presence of the object class for any of the selected anchor boxes. In many popular detectors \cite{FasterRCNN, SSD, RetinaNet, RefineDet, M2Det}, the selection of these anchors is performed by simple thresholding of IOU between the default anchor box and the ground truth box, as shown in Table \ref{table:anchor-method}. 

An apparent issue with such binary thresholding is that only a slight difference in IOU leads to opposite results; for instance, an anchor with IOU$=0.501$ is chosen as a positive sample, whereas that with IOU$=0.499$ is not. Our conjecture is that this coarse selection of samples leads to the aforementioned improper behaviors of detectors. 


\begin{table}[tb]
    \caption{Various strategies to select positive (and negative) samples in popular detectors, which are either from the original papers or from authors' implementations. 
    $C_{ij}$ in YOLOv2 is an indicator showing if the center of an object is on the grid $i$ and its associated anchor $j$ achieves the highest IOU with the ground-truth boundary box.
    }
    \label{table:anchor-method}
    \begin{center}
    \scalebox{0.7}{
        \begin{tabular}{c|cccc}
            \hline
            \multicolumn{1}{c|}{models} & 
            \multicolumn{1}{c}{positive} & 
            \multicolumn{1}{c}{negative} & 
            \multicolumn{1}{c}{anchors} &
            \multicolumn{1}{c}{HNM} \\
            \hline\hline
            
            \multicolumn{1}{c|}{Faster R-CNN \cite{FasterRCNN}} &
            \multicolumn{1}{c}{IOU $> 0.7$} & 
            \multicolumn{1}{c}{IOU $< 0.3$} & 
            \multicolumn{1}{c}{9} &
            \multicolumn{1}{c}{ - } \\
    
            \multicolumn{1}{c|}{SSD \cite{SSD}} & 
            \multicolumn{1}{c}{IOU $> 0.5$} & 
            \multicolumn{1}{c}{ - } & 
            \multicolumn{1}{c}{4 or 6} &
            \multicolumn{1}{c}{1:3} \\
            
            \multicolumn{1}{c|}{RetinaNet \cite{RetinaNet}} & 
            \multicolumn{1}{c}{IOU $\ge 0.5$} & 
            \multicolumn{1}{c}{IOU $< 0.4$} & 
            \multicolumn{1}{c}{9} &
            \multicolumn{1}{c}{ - } \\
            
            \multicolumn{1}{c|}{RefineDet \cite{RefineDet}} & 
            \multicolumn{1}{c}{IOU $> 0.5$} & 
            \multicolumn{1}{c}{ - } & 
            \multicolumn{1}{c}{4} &
            \multicolumn{1}{c}{1:3} \\
            
            \multicolumn{1}{c|}{M2Det \cite{M2Det}} & 
            \multicolumn{1}{c}{IOU $\ge 0.5$} & 
            \multicolumn{1}{c}{ - } & 
            \multicolumn{1}{c}{6} &
            \multicolumn{1}{c}{1:3} \\
            \hline
    
            \multicolumn{1}{c|}{YOLOv2 \cite{yolov2}} & 
            \multicolumn{1}{c}{$C_{ij}=1$} & 
            \multicolumn{1}{c}{$C_{ij}=0$\ and\ IOU $\le 0.6$} & 
            \multicolumn{1}{c}{5} &
            \multicolumn{1}{c}{ - } \\
            \hline
        \end{tabular}
    }
    \end{center}
\end{table}

\subsubsection{Proposed Sampling Method}\label{sec:ASmmd.improve.propose}

We propose to make a softer decision for the selection of positive samples. The basic idea is to incorporate a continuous weight tied with the IOU between the anchor box and the ground truth box into the evaluation of the loss. In a standard implementation of the conventional sampling scheme, an indicator variable $X_k\in\{0,1\}$ is used to express whether the $k$-th anchor box will be chosen as a positive sample or not, which is computed by thresholding the IOU, and then it is used as a weight of the loss for this anchor box. We extend $X_k$ to the continuous domain, i.e.,  $X_k'\in[0,1]$;  $X_k'$ is multiplied with the loss for the $k$-th anchor as in the conventional method. 

We compute the new indicator $X_k'$ using the IOU $r$ between the anchor $k$ and the ground truth bounding box as
\begin{equation}
    X_k'=f(r),
\end{equation}
where $f$ is a logistic sigmoid function with a parameter $a$:
\begin{equation}
    f(r) = \frac{1}{1+\exp\{-a(r-0.5)\}}.
\end{equation}
For the sake of computational efficiency, we consider this soft thresholding only in the range $[0.5-\alpha,0.5+\alpha]$ of IOU $r$. 
That is, we simply set $X_k'=0$ for $r<0.5-\alpha$ and $X_k'=1$ for $r>0.5+\alpha$. 
To smoothly connect $X_k'$ at the borders $r=0.5\pm\alpha$, we set $a$ so that $f(0.5-\alpha)=\beta$ (or equivalently, $f(0.5+\alpha)=1-\beta$) for a small $\beta$. We set $\alpha=0.1$ and $\beta=0.001$ throughout our experiments. To avoid the case where no anchor is assigned to a ground truth box, we follow previous studies, setting $X_k'=1$ for the anchor $k$ with the largest IOU, when no anchor with $X_k>0$ exists. 

\begin{figure}[t]
    \begin{center}
    \scalebox{0.95}{
        \includegraphics[keepaspectratio, width=\linewidth]{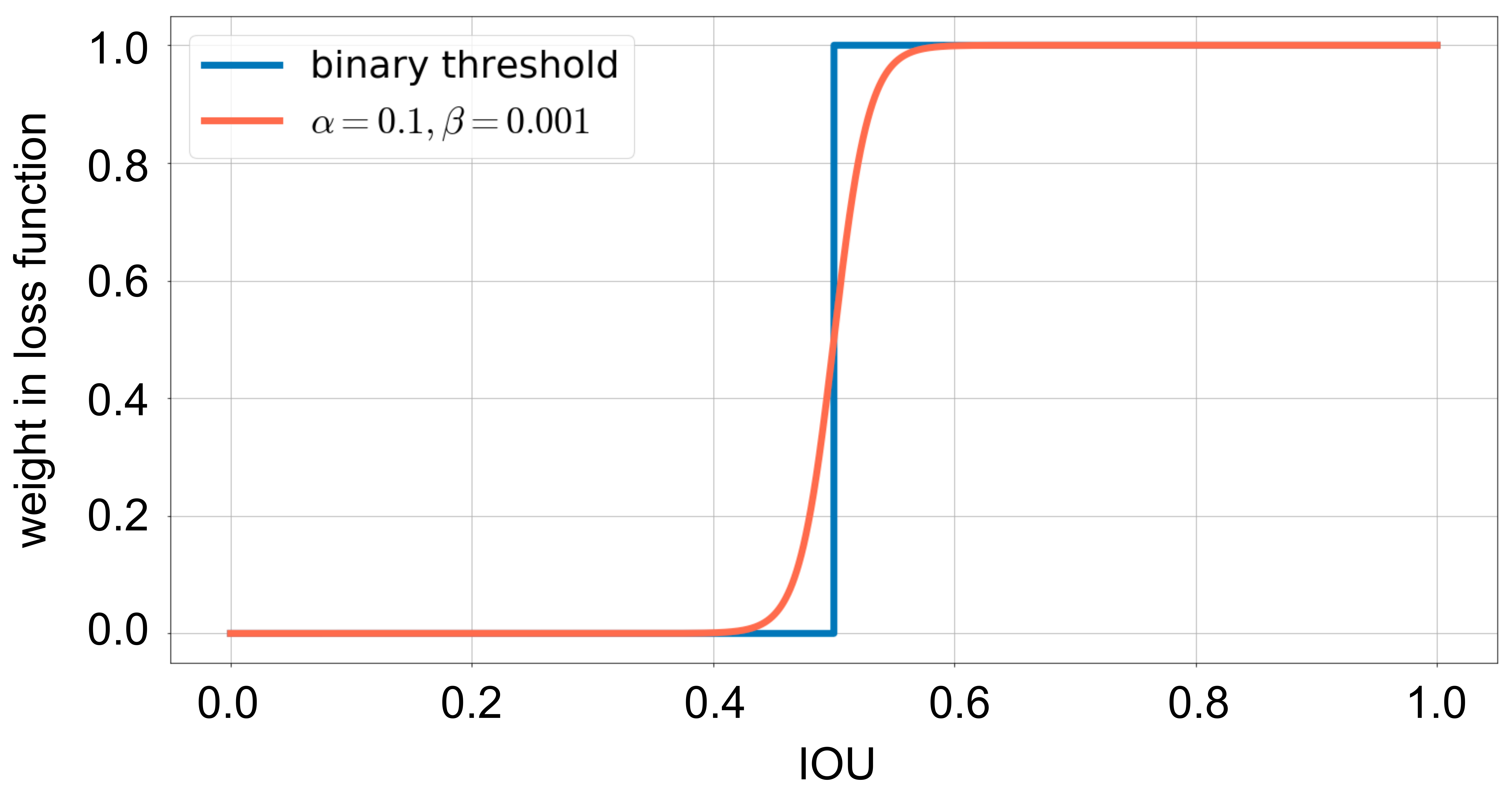}
        }
    \end{center}
    \caption{Incorporated weight $X_k'$ applied on the loss of anchor $k$ having IOU $r$ with the ground truth box ($\alpha=0.1$ and $\beta=0.001$). Many previous methods employ binary thresholding with IOU $=0.5$ to choose positive samples. Our weight performs soft thresholding. }
    \label{fig:threshold}
\end{figure}

\color{black}

\section{Experimental Results}\label{sec:result}


\subsection{Detectors}\label{sec:result.detector}

For detectors, we choose SSD \cite{SSD} and M2Det \cite{M2Det}, which are popular two detectors
that are based on anchor boxes. For the sake of experimental reproducibility, we used their implementation that are publicly available, i.e., SSD-VGG16\footnote{https://github.com/qfgaohao/pytorch-ssd}, SSD-ResNet50\footnote{https://github.com/ShuangXieIrene/ssds.pytorch}, and M2Det-VGG16\footnote{https://github.com/qijiezhao/M2Det} (VGG16/ResNet50 indicates a backbone network \cite{VGG, ResNet}). We trained them on the most popular dataset for object detection, PASCAL VOC. 

\paragraph{Details of training} The PASCAL VOC dataset \cite{VOC} $2007+2012$ trainval split is used for training. SGD with momentum was used for the optimizer, where momentum was set to $0.9$ with weight decay $5.0 \times 10^{-4}$ in all models. In the training of SSD \cite{SSD}, setting the initial learning rate to $1\times10^{-3}$, we decrease it to $1\times10^{-4}$  at 120 epochs and  $1\times10^{-5}$  at 160 epochs; training is stopped at 200 epochs.
In the training of M2Det \cite{M2Det}, initial learning rate is set to  $4\times10^{-3}$, it is decreased to $2\times10^{-3}$, $4\times10^{-4}$, $4\times10^{-5}$, and $4\times10^{-6}$ at 40, 60, 80, and 100 epochs; training is stopped at 120 epochs. For M2Det, weights are initialized by a model pretrained on COCO dataset \cite{MSCOCO} and finetune them.



\subsection{Dataset}\label{sec:result.dataset}

We choose DAVIS (Densely Annotated VIdeo Segmentation)  2017 dataset \cite{DAVIS} for our experiments. Existing datasets for object detection such as PASCAL VOC \cite{VOC} and COCO \cite{MSCOCO} contain only still-images and cannot be used. Datasets for object tracking such as \cite{OTB, MOT} do contain videos but tend to lack a variety of objects and changes in object size, aspect ratio, etc. The DAVIS dataset is ideal for our purpose, since it consists of 90 videos of various scenes and pixel-wise segmentation masks are provided for various objects in each frame of all the video. We automatically generate ground truth bounding boxes by obtaining the circumscribing box to each segmentation mask of an object. 


As they are trained on PASCAL VOC, our detectors are trained to detect the 20 classes of objects contained in the dataset. We identify 14\footnote{aeroplane, bicycle, bird, boat, bus, car, cat, cow, dog, horse, motorbike, person, sheep, and train} of them in the videos of the DAVIS dataset, and thus use these objects for our experiments. We can use only 73 videos out of 90 and discard others that do not contain the 20 object classes. As there are often multiple objects (that are annotated) in an image and our analysis is done for each object, the number of video frames we used in our experiments amounts to $8,140$.


\paragraph{Extraction of MMD Frames}

We run each of the above detectors on the above data and then applied the criteria (\ref{eq:mmd_condition}) to the detection results to extract target frames for which MMD occurs. This yields several hundreds of frames for each detector; exact numbers are shown in the row of `MMD frames' of Table \ref{table:eval_result}. It is seen that the these frames are less than 5\% of the total number of frames (i.e., $8,140$).

\subsection{Classifying MMD Frames by Causes}\label{sec:result.cls}
\begin{table}[tb]
    \begin{center}
        \caption{Number of MMD frames found on selected video frames (8,140 images in total from 73 videos) of the DAVIS dataset for different detectors, and their classification into three categories. $^\dagger$mAP for each detector is evaluated on PASCAL VOC 2007 test set.}    
    \label{table:eval_result}
    \medskip
    \scalebox{0.85}{
    \begin{tabular}{c|cccc}
        \hline
        \multicolumn{1}{c|}{Models} & 
        \multicolumn{1}{c}{SSD{\footnotesize -VGG16}} & 
        \multicolumn{1}{c}{SSD{\footnotesize -ResNet50}} & 
        \multicolumn{1}{c}{M2Det{\footnotesize -VGG16}} \\ 
        \hline\hline
        
        
        \multicolumn{1}{c|}{mAP$_{0.5}^\dagger$} & 
        \multicolumn{1}{c}{76.5} & 
        \multicolumn{1}{c}{74.8} & 
        \multicolumn{1}{c}{79.3} \\
        \hline
        
        \multicolumn{1}{c|}{MMD frames} & 
        \multicolumn{1}{c}{367} &
        \multicolumn{1}{c}{333} & 
        \multicolumn{1}{c}{264} \\
        \hline
        
        \multicolumn{1}{c|}{External factors} &
        \multicolumn{1}{c}{262} &
        \multicolumn{1}{c}{239} & 
        \multicolumn{1}{c}{160} \\
        
        \multicolumn{1}{c|}{Anchor boundary} & 
        \multicolumn{1}{c}{\bf 73} &
        \multicolumn{1}{c}{\bf 61} & 
        \multicolumn{1}{c}{\bf 75} \\
        
        \multicolumn{1}{c|}{Others} & 
        \multicolumn{1}{c}{32} &
        \multicolumn{1}{c}{33} & 
        \multicolumn{1}{c}{29} \\
        \hline
    \end{tabular}
    }
    \end{center}
\end{table}

We then visually inspect each of these MMD frames. We first check if the MMD at the frame is caused by an external factor, such as motion blur, occlusion, and cluttered background, as discussed in Sec.~\ref{sec:mmd.pf.ext}. There are frames at which the object is correctly detected but its predicted class is wrong, e.g., a dog is detected but is recognized as a cat; there are about 5 to 10 frames in total for each detector. We classified these frames also into the category of external factors. The row `External factors' of Table \ref{table:eval_result} shows the number of these frames for each detector. 

We next inspect each of the remaining MMD frames to judge whether it emerges due to the {\em anchor boundary} as is predicted in Hypothesis 1 or due to {\em other causes}. To do this, we apply the method in Sec.~\ref{sec:ASmmd} to these MMD frames; this yields sequences of warped images, for each of which we run the same detector. We then plot detection scores outputted for the associated neighboring anchors, by using which we make the above decision. The rows `anchor boundary' and `others' of Table \ref{table:eval_result} show the number of MMD frames classified into the two categories, respectively.

\begin{figure}[t]
    \begin{center}
        \includegraphics[keepaspectratio, width=\linewidth]{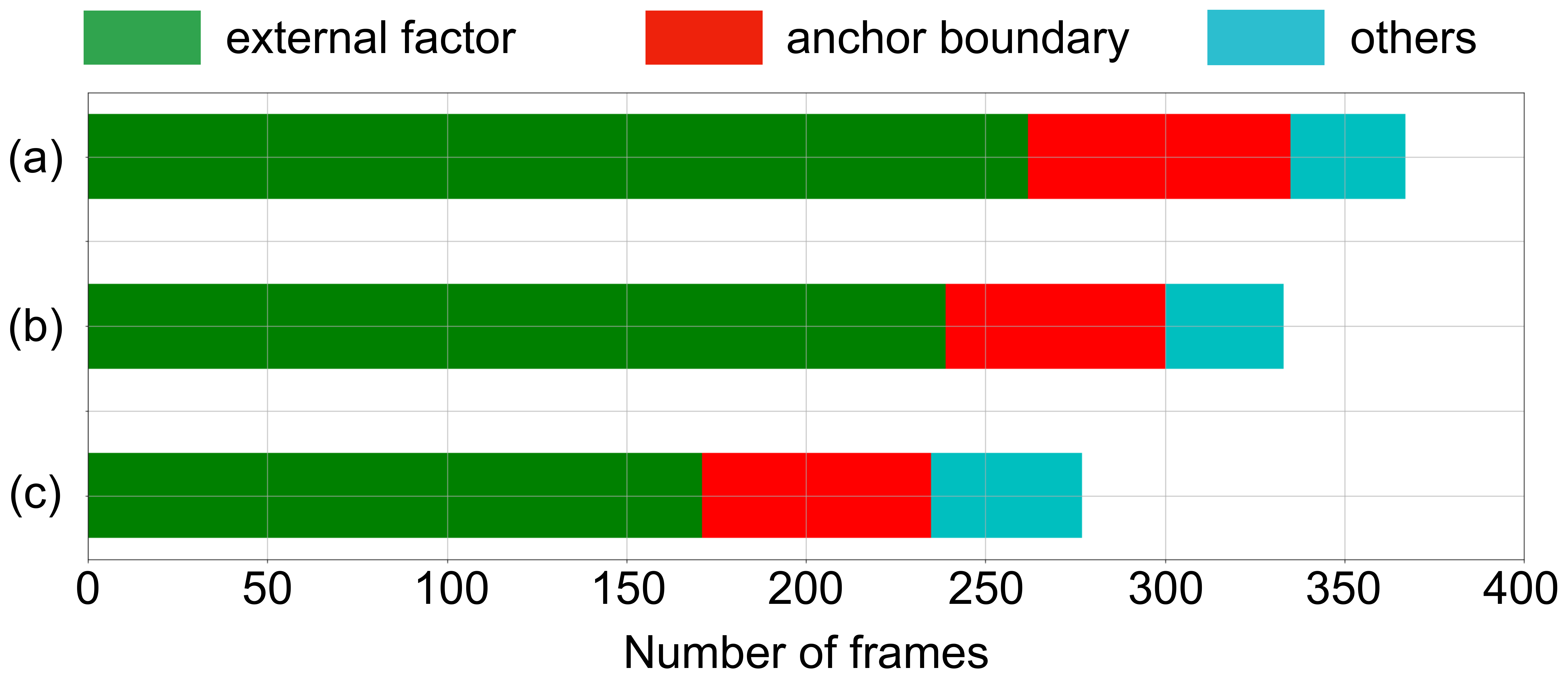}\\
    \end{center}
    \caption{Bar plots of the same data shown in Table \ref{table:eval_result}. (a) SSD-VGG16, (b) SSD-ResNet50, and (c) M2Det-VGG16. 
    }
    \label{fig:eval_result}
\end{figure}

These results are summarized in the bar plots of Fig.~\ref{fig:eval_result}. It is seen that while the external factors are dominant, the hypothesized cause of anchor boundaries occupies non-negligible portion. It will be the most effective if we can cope with the external factors; for instance, it may be effective to remove motion blur from the input image in a pre-processing step before the application of an object detector. This is, however, not a simple task. On the other hand, as will be shown in the next subsection, we can reduce the number of MMDs caused by anchor boundaries {\em for free}, i.e., by simply switching the method for positive sample selection from the conventional one to the proposed method. 

{
}

Note that there remain a portion of MMD frames that are not explained by the external factors or the anchor boundaries. It should also be noted that we could not find a MMD frame caused by anchor boundaries in terms of aspect ratio. This may be because objects tend to change their aspect ratio less frequently than they change their sizes or exhibit translational motion. 

{

}

\subsection{Evaluation of Proposed Sampling Method}\label{sec:result.eval}

\begin{figure}[t]
    \begin{center}
    \scalebox{1.0}{
        \includegraphics[keepaspectratio, width=\linewidth]{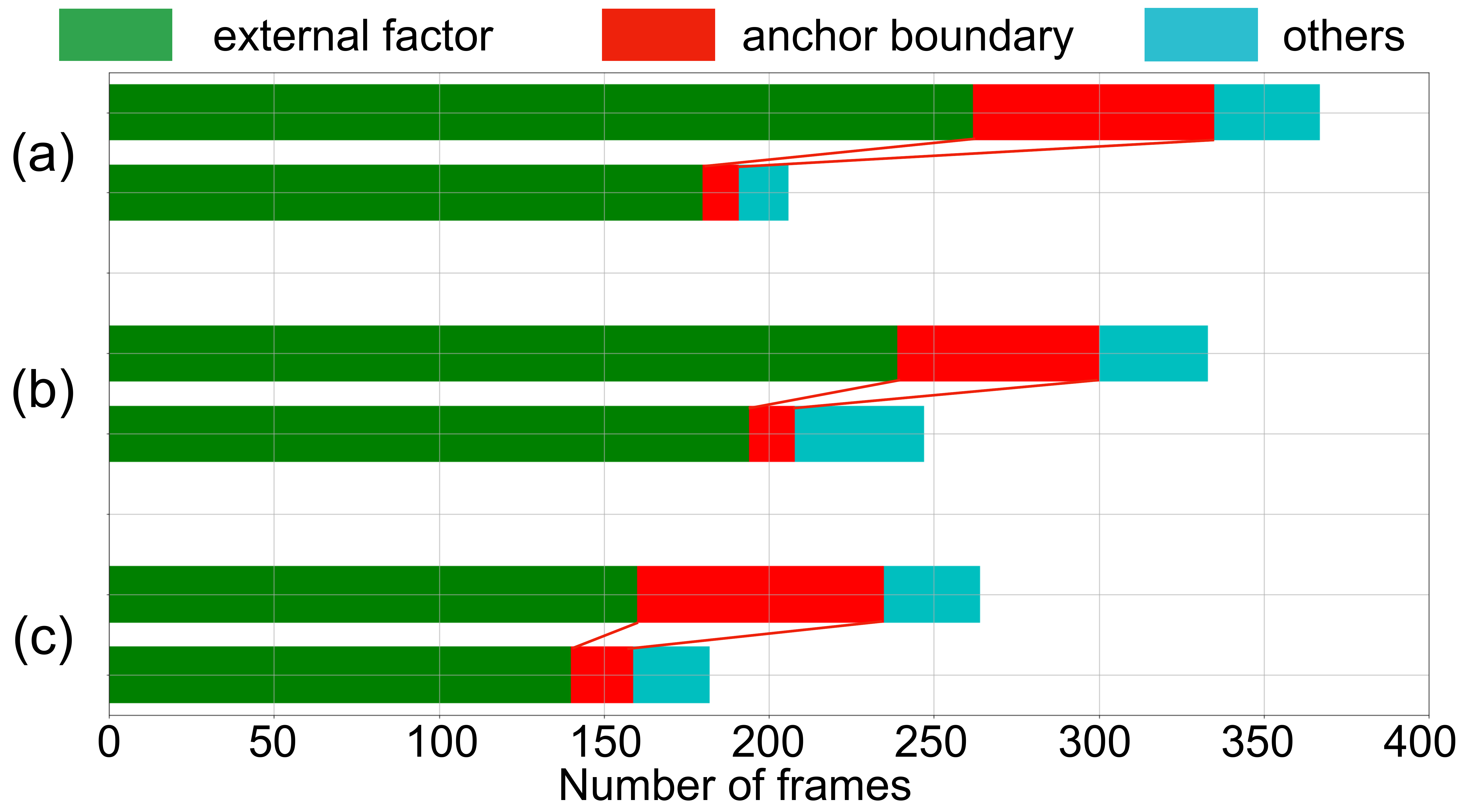}
    }
    \end{center}
    \caption{Bar plots of  Table~\ref{table:sigmoid_result} (and Table ~\ref{table:eval_result} for the sake of comparison). (a) SSD-VGG16, (b) SSD-ResNet50, and (c) M2Det-VGG16. It is seen that the number of MMD frames, particularly those due to anchor boundaries, decreases.} 
    \label{fig:improved_result}
\end{figure}

We also conducted experiments to examine the effectiveness of the proposed method for selecting positive samples and weighting them. In the experiments we trained the same three detector models, SSD-VGG16, SSD-ResNet50, and M2Det-VGG16 using the proposed sampling/weighting method explained in Sec.~\ref{sec:ASmmd.improve}. Their training in other parts is performed in the same way as before. We applied the trained models on the same dataset (i.e., 8,140 images from 73 DAVIS videos) and then  analyzed the detection results using the same method as the previous experiments. The results are shown in Table \ref{table:sigmoid_result}. The bar plot of the same data including Table \ref{table:eval_result} are shown in Fig.~\ref{fig:improved_result}. 

It is seen from the table and the figure that the employment of the proposed sampling/weighting method decreases the number of MMD frames caused by anchor boundaries significantly; namely, from 73 to 11 with SSD-VGG16, from 61 to 14 with SSD-ResNet50, and from 75 to 19 with M2Det-VGG16. 

\begin{figure*}[t]
    \begin{center}
    \scalebox{0.95}{
        \includegraphics[keepaspectratio, width=\linewidth]{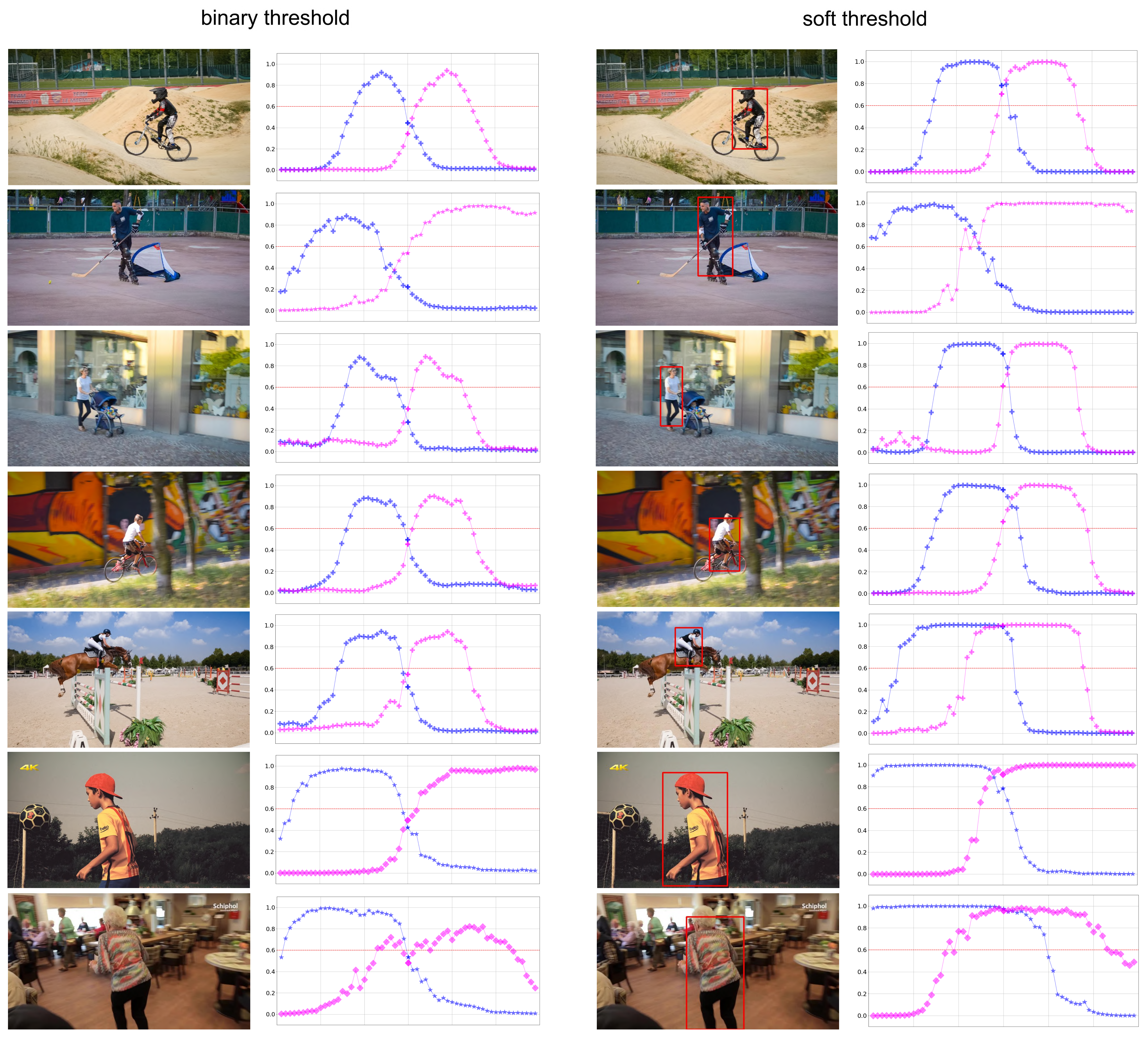}
    }
    \end{center}
    \caption{Examples of frames at which SSD-VGG16 trained with the conventional method for positive sample selection (i.e., binary thresholding) causes MMD, whereas the same model trained with the proposed method (i.e., soft thresholding) ceases to cause MMD. The plots on the second columns show the scores for two neighboring anchors, from which it is confirmed that the proposed sampling method contributes to fill in the gap at around anchor boundaries.}
    
    \label{fig:restored_images}
\end{figure*}

Figure~\ref{fig:restored_images} shows a number of examples of how MMD is resolved by the incorporation of the proposed method. The left panel shows detection results of SSD-VGG16 with the conventional sampling method (i.e., binary thresholding), which suffer from MMD. Each row shows a pair of a MMD frame and the score profile for neighboring anchors over associated image warping. As is explained in Sec.~\ref{sec:ASmmd.visualize.case_s}, it is seen for each case that there emerges a valley in the score profile at around the anchor boundary, which well explains why MMD occurs at the frame. The right panel shows results on the same frames of the same detector trained with the proposed method. Each row shows a pair of the detection result and the corresponding score profiles for the neighboring anchors. In each case, the object that is miss-detected on the left panel is correctly detected. It is seen for each case that the peak of the score curve for each anchor tends to be more flat, resulting in shallower valley at the anchor boundary. This well explains the vanishing of MMDs in these examples. 

We may conclude from these results the effectiveness of our approach. Moreover, this will also be an additional support for the validity of Hypothesis 1. Another remark with the results is that MMD frames caused by external factors also decreases to a certain degree; the largest decrease (more than 30\%) is observed for SSD-VGG16. Although the mechanism behind this improvement is not clear, we can confirm the proposed method does not worsen the performance or even contribute to improvements overall. This is also confirmed by that mAP maintains the same level or even shows some improvements. As is mentioned earlier, we set $\alpha=0.1$ and $\beta=0.001$ in the experiments. It is noteworthy that we chose these values intuitively and there may be better values leading to even better performance. 

{
}

{


}

\begin{table}[tb]
    \caption{Comparisons between binary and soft thresholding for positive sample selection. 
    We set hyperparameters for soft threshold as [$\alpha$, $\beta$]=[0.1, 0.001].
    $^\dagger$mAP values are evaluated on the PASCAL VOC 2007 test set.}
    \label{table:sigmoid_result}
    \begin{center}
    \scalebox{0.80}
    {
        \begin{tabular}{c|cc|cc|cc}
            \hline
            \multicolumn{1}{c|}{Models} & 
            \multicolumn{2}{c|}{SSD{\footnotesize -VGG16}} &
            \multicolumn{2}{c|}{SSD{\footnotesize -ResNet50}} & 
            \multicolumn{2}{c}{M2Det{\footnotesize -VGG16}} \\  
            \hline
            
            \multicolumn{1}{c|}{Threshold} & 
            \multicolumn{1}{c}{Binary} & 
            \multicolumn{1}{c|}{Soft} &
            \multicolumn{1}{c}{Binary} & 
            \multicolumn{1}{c|}{Soft} & 
            \multicolumn{1}{c}{Binary} & 
            \multicolumn{1}{c}{Soft} \\
            \hline\hline
            
            \multicolumn{1}{c|}{mAP$_{0.5}^\dagger$} & 
            \multicolumn{1}{c}{76.5} & 
            \multicolumn{1}{c|}{77.2} & 
            \multicolumn{1}{c}{74.8} &
            \multicolumn{1}{c|}{74.7} &
            \multicolumn{1}{c}{79.3} &
            \multicolumn{1}{c}{80.4} \\
            \hline
            
            \multicolumn{1}{c|}{MMD frames} & 
            \multicolumn{1}{c}{367} &
            \multicolumn{1}{c|}{206} &
            \multicolumn{1}{c}{333} &
            \multicolumn{1}{c|}{247} &
            \multicolumn{1}{c}{264} &
            \multicolumn{1}{c}{182} \\
            \hline
            
            \multicolumn{1}{c|}{External factors} &
            \multicolumn{1}{c}{262} &
            \multicolumn{1}{c|}{180} &
            \multicolumn{1}{c}{239} &
            \multicolumn{1}{c|}{194} &
            \multicolumn{1}{c}{160} &
            \multicolumn{1}{c}{140} \\
            
            \multicolumn{1}{c|}{Anchor boundary} & 
            \multicolumn{1}{c}{73} &
            \multicolumn{1}{c|}{\textbf{11}} &
            \multicolumn{1}{c}{61} &
            \multicolumn{1}{c|}{\textbf{14}} &
            \multicolumn{1}{c}{75} &
            \multicolumn{1}{c}{\textbf{19}} \\
            
            \multicolumn{1}{c|}{Others} & 
            \multicolumn{1}{c}{32} &
            \multicolumn{1}{c|}{15} &
            \multicolumn{1}{c}{33} &
            \multicolumn{1}{c|}{39} &
            \multicolumn{1}{c}{29} &
            \multicolumn{1}{c}{23} \\
            \hline
        \end{tabular}
    }
    \end{center}
\end{table}

\section{Summary and Conclusion}\label{sec:conclusion}

Recent CNN-based detectors are designed to work on a single image and are usually trained using a large number of labeled still-images that are independent of each other. In this paper, we have analyzed momentarily missed detection (MMD) that is often observed when we apply these detectors to a sequence of video frames. We have revealed through several experiments that i) external factors (e.g., motion blur etc.) explains the majority of MMD cases; ii) the remaining MMD cases can be mostly explained by an improper behavior of the detectors at boundaries of anchor boxes; and iii) it can be rectified by using the improved method that chooses positive samples from candidate anchor boxes when training the detectors. 



\subsection*{Acknowledgements} \label{sec:acknowledgements}
This work was partly supported by JSPS KAKENHI Grant Number JP15H05919 and JP19H01110 and JST CREST Grant Number JPMJCR14D1.

{\small
\bibliographystyle{ieee}
\bibliography{my_egbib}

\begin{thebibliography}{10}\itemsep=-1pt

\bibitem{CenterNet}
K.~Duan, S.~Bai, L.~Xie, H.~Qi, Q.~Huang, and Q.~Tian.
\newblock Centernet: Keypoint triplets for object detection.
\newblock In {\em Proc. CVPR}, 2019.

\bibitem{VOC}
M.~Everingham, L.~V.~Gool, C.~K.~I. Williams, J.~Winn, and A.~Zisserman.
\newblock {The Pascal Visual Object Classes (VOC) Challenge}.
\newblock {\em IJCV}, 2010.

\bibitem{DSSD}
C.~Y. Fu, W.~Liu, A.~Ranga, A.~Tyagi, and A.~C. Berg.
\newblock {DSSD} : Deconvolutional single shot detector.
\newblock {\em arXiv:1701.06659}, 2017.

\bibitem{ResNet}
K.~He, X.~Zhang, S.~Ren, and J.~Sun.
\newblock {Deep Residual Learning for Image Recognition}.
\newblock In {\em Proc. CVPR}, 2016.

\bibitem{DenseBox}
L.~Huang, Y.~Yang, Y.~Deng, and Y.~Yu.
\newblock Densebox: Unifying landmark localization with end to end object
  detection.
\newblock {\em arXiv:1509.04874}, 2015.

\bibitem{CornerNet}
H.~Law and J.~Deng.
\newblock Cornernet: Detecting objects as paired keypoints.
\newblock In {\em Proc. ECCV}, 2018.

\bibitem{MSCOCO}
T.~Lin, M.~Maire, S.~J. Belongie, L.~D. Bourdev, R.~B. Girshick, J.~Hays,
  P.~Perona, D.~Ramanan, P.~Doll{\'{a}}r, and C.~L. Zitnick.
\newblock Microsoft {COCO:} common objects in context.
\newblock In {\em Proc. ECCV}, 2014.

\bibitem{FPN}
T.~Y. Lin, P.~Dollar, R.~Girshick, K.~He, B.~Hariharan, and S.~Belongie.
\newblock {Feature Pyramid Networks for Object Detection}.
\newblock In {\em Proc. CVPR}, 2017.

\bibitem{RetinaNet}
T.-Y. Lin, P.~Goyal, R.~Girshick, K.~He, and P.~Dollar.
\newblock Focal loss for dense object detection.
\newblock In {\em Proc. ICCV}, 2017.

\bibitem{RTR}
C.~{Liu}, P.~{Liu}, W.~{Zhao}, and X.~{Tang}.
\newblock Robust tracking and redetection: Collaboratively modeling the target
  and its context.
\newblock {\em TMM}, 2018.

\bibitem{SSD}
W.~Liu, D.~Anguelov, D.~Erhan, C.~Szegedy, S.~Reed, C.~Fu, and A.~C. Berg.
\newblock {SSD: Single Shot MultiBox Detector}.
\newblock In {\em Proc. ECCV}, 2016.

\bibitem{UnderMB}
B.~{Ma}, L.~{Huang}, J.~{Shen}, L.~{Shao}, M.~{Yang}, and F.~{Porikli}.
\newblock Visual tracking under motion blur.
\newblock {\em TIP}, 2016.

\bibitem{MOT}
A.~Milan, L.~Leal{-}Taix{\'{e}}, I.~D. Reid, S.~Roth, and K.~Schindler.
\newblock {MOT16:} {A} benchmark for multi-object tracking.
\newblock {\em arXiv:1603.00831}, 2016.

\bibitem{GS_Face}
X.~Ming, F.~Wei, T.~Zhang, D.~Chen, and F.~Wen.
\newblock Group sampling for scale invariant face detection.
\newblock In {\em Proc. CVPR}, 2019.

\bibitem{DAVIS}
J.~Pont{-}Tuset, F.~Perazzi, S.~Caelles, P.~Arbelaez, A.~Sorkine{-}Hornung, and
  L.~V. Gool.
\newblock {The 2017 {DAVIS} Challenge on Video Object Segmentation}.
\newblock {\em arXiv:1704.00675}, 2017.

\bibitem{yolov2}
J.~Redmon and A.~Farhadi.
\newblock Yolo9000: Better, faster, stronger.
\newblock In {\em Proc. CVPR}, 2017.

\bibitem{FasterRCNN}
S.~Ren, K.~He, R.~Girshick, and J.~Sun.
\newblock {Faster R-CNN: Towards Real-Time Object Detection with Region
  Proposal Networks}.
\newblock In {\em Proc. NIPS}, 2015.

\bibitem{VGG}
K.~Simonyan and A.~Zisserman.
\newblock {Very Deep Convolutional Networks for Large-Scale Image Recognition}.
\newblock In {\em Proc. ICLR}, 2015.

\bibitem{FCOS}
Z.~Tian, C.~Shen, H.~Chen, and T.~He.
\newblock {FCOS:} fully convolutional one-stage object detection.
\newblock In {\em Proc. ICCV}, 2019.

\bibitem{ResMap}
R.~Walsh and H.~Medeiros.
\newblock Detecting tracking failures from correlation response maps.
\newblock In {\em Proc. ISVC}, 2016.

\bibitem{OTB}
Y.~Wu, J.~Lim, and M.-H. Yang.
\newblock Object tracking benchmark.
\newblock {\em TPAMI}, 2015.

\bibitem{STMN}
F.~Xiao and Y.~J.~Lee.
\newblock {Video Object Detection with an Aligned Spatial-Temporal Memory}.
\newblock In {\em Proc. ECCV}, 2018.

\bibitem{Metaanchor}
T.~Yang, X.~Zhang, Z.~Li, W.~Zhang, and J.~Sun.
\newblock Metaanchor: Learning to detect objects with customized anchors.
\newblock In {\em Proc. NIPS}, 2018.

\bibitem{DetStability}
H.~Zhang and N.~Wang.
\newblock {On The Stability of Video Detection and Tracking}.
\newblock {\em arXiv:1611.06467}, 2016.

\bibitem{RefineDet}
S.~Zhang, L.~Wen, X.~Bian, Z.~Lei, and S.~Z. Li.
\newblock {Single-Shot Refinement Neural Network for Object Detection}.
\newblock In {\em Proc. CVPR}, 2018.

\bibitem{M2Det}
Q.~Zhao, T.~Sheng, Y.~Wang, Z.~Tang, Y.~Chen, L.~Cai, and H.~Ling.
\newblock M2det: A single-shot object detector based on multi-level feature
  pyramid network.
\newblock In {\em Proc. AAAI}, 2019.

\bibitem{FeatureSelective}
C.~Zhu, Y.~He, and M.~Savvides.
\newblock Feature selective anchor-free module for single-shot object
  detection.
\newblock In {\em Proc. CVPR}, 2019.

\bibitem{Flowguided}
X.~Zhu, Y.~Wang, J.~Dai, L.~Yuan, and Y.~Wei.
\newblock {Flow-Guided Feature Aggregation for Video Object Detection}.
\newblock In {\em Proc. ICCV}, 2017.

\end{thebibliography}
}

\end{document}